%% file: main.tex
\PassOptionsToPackage{nosubfloats}{jmlrutils}

\documentclass{template}
\usepackage{graphicx}
\usepackage{booktabs,array}
\usepackage{tikz}
\usetikzlibrary{arrows.meta,decorations.markings,calc}
\usepackage{cuted}
\usepackage{caption}
\usepackage{subcaption}
\usepackage{wrapfig}
\usepackage{ragged2e}

\newtheorem{assumption}{Assumption}
\newcommand{\qed}{\hfill$\Box$}
\RequirePackage[T1]{fontenc}

\usepackage{etoc}

\newcommand{\trace}{\operatorname{tr}}

\AtBeginDocument{%
  \setlength{\parindent}{0pt}
  \setlength{\parskip}{0.5\baselineskip}
}

\AtBeginDocument{%

  \setcounter{topnumber}{4}
  \setcounter{bottomnumber}{2}
  \setcounter{totalnumber}{6}
}

\makeatletter
\AtBeginDocument{%
  \jmlrproceedings{}{}%
  \jmlrvolume{}%
  \jmlrissue{}%
  \jmlryear{}%
  \jmlrpages{}%
  \jmlrworkshop{}%
  \jmlrsubmitted{}%
  \jmlrpublished{}%
  \editor{}%

  \let\@titlefoot\@empty
  \let\ps@jmlrtps\ps@plain
}
\makeatother

\title{Exact Scaling Laws for a Random-Features Bigram Model}

\author{%
Julian Enoch$^{1,2}$\\
Elvis Dohmatob$^{1,2}$\\
    \addr{$^1$Concordia University}\\
  \addr{$^2$Mila -- Quebec AI Institute}
}

\usepackage{color}

\begin{document}
\etocdepthtag.toc{main}

%

%

\maketitle

\begin{abstract}
Understanding neural scaling laws is increasingly important for theory and practice, yet first-principles explanations remain limited even in simple settings. In this work, we study capacity-limited scaling in a bottlenecked bigram model with power-law (Zipf) token statistics, viewed as an analytically tractable toy model of a large language model. Fixed random embeddings serve as a proxy for representational capacity and induce an explicit bottleneck of width, allowing us to derive scaling laws for prediction regret as a function of model width, vocabulary size, and the Zipf exponent. Our results complement recent theoretical accounts based on optimization dynamics by isolating scaling due purely to representational constraints. In the appropriate regime, the theory recovers the functional form of classical neural scaling laws and suggests that heavy-tailed language statistics may underlie both optimization- and capacity-driven scaling. Experiments natural language datasets (WikiText2, OpenWebText) closely match the predicted curves.
\end{abstract}

\input{intro}

\section{Theoretical Setup}
Like \citet{kunstner2025scaling}, we consider a bigram model in which each word is associated with a unique integer $i \in [d]$, where $[d] := \{1,2,\ldots,d\}$ is the vocabulary, and $d \gg $1 is the vocabulary size. The frequency of $i$th token/word is $p_i$, and conditioned on $k$ the frequency of the next word being $j$ is $\pi_{j \mid i}$. This defines a distribution $P$ over word pairs. We make the following assumption on this distribution.
\begin{assumption}(i) \textbf{Zipf marginals.} There exists an exponent $\alpha > 0$ such that the $i$th word has frequency
\begin{eqnarray}
\label{eq:Zipf}
p_i = \frac{i^{-\alpha}}{H_{d,\alpha}},\text{ with }H_{d,\alpha}:=\sum_{k=1}^d \frac{1}{k^\alpha}.
\end{eqnarray}
 
(ii) \textbf{Constant collision probability.} There exists a constant $A \in (0,1]$ such that for each word $i \in [d]$,
\begin{eqnarray}
    \sum_{j=1}^d \pi_{j \mid i}^2 = A.
    \label{eq:A}
\end{eqnarray}
\label{ass:main}
\end{assumption}
The distributional constants $\alpha$ and $A$ will play a crucial role in our theory. A similar assumption was made in \citep{cabannes2024associative}. Also,  \citet{kunstner2025scaling} considered a similar but strong similar assumption as above was also made in and \citep{kunstner2025scaling}. Indeed, in place of the second part of the above assumption, they assumed that the next-word frequencies also followed a Zipf law with exponent $\alpha'>0$. Such distributions verify the invariance \eqref{eq:A} above with $A=H_{d,2\alpha'}/H_{d,\alpha'}^2$.

\begin{table*}[!htp]
    \centering
    \begin{tabular}{lccc}
        \toprule
        \textbf{Dataset} & \textbf{Vocabulary size} $d$ & \textbf{Sample size} $n$ & \textbf{Zipf exponent} $\alpha$\\
        \midrule
WikiText-2 & 33277 & 2051910 & 0.9354\\
OpenWebText & 50257 & 2110824721 & 0.9443\\
        \bottomrule
    \end{tabular}
    \caption{Global parameters and descriptive statistics for real-world natural language datasets. In our experiments, we sample subsets of these datasets, restricting both the vocabulary size $d$ and the sample size $n$.}
    \label{tab:params}
\end{table*}

\subsection{Unconstrained Bigram Model}
Our proposed capacity-constrained bigram model (Section \ref{sec:RF-bigram}) is motivated by the following full-capacity bigram model considered by \citet{kunstner2025scaling}, which we now describe. 

Nature draws an i.i.d.\ sample $\{(x_1,y_1),\ldots,(x_n,y_n)\}$
of size $n$ from $P$, where each $x_t$ and $y_t$ is the one-hot
encoding of the corresponding symbol in the alphabet $[d]$. Table \ref{tab:params} lists the values for the sample size $n$ and the vocabulary size $n$ for popular natural language datasets: WikiText-2 \citep{wikitext} and OpenWebText2 \citep{openwebtext}. Let $X$ and $Y$ be the $n\times d$ binary matrices whose $t$th
rows are $x_t^\top$ and $y_t^\top$, respectively. \citet{kunstner2025scaling} considered the problem of approximating
$Y$ by $XW$ through minimization of the loss
\begin{equation}
    L_{n,d}(W)
    := n^{-1}\|XW-Y\|_{\mathrm{Fro}}^2,
    \label{eq:L}
\end{equation}
over a parameter matrix $W\in\mathbb{R}^{d\times d}$.
Here, $\|Z\|_{\mathrm{Fro}}^2
:=\operatorname{tr}Z^\top Z$ denotes the squared Frobenius
(aka Hilbert--Schmidt) norm of a matrix $Z$.
The learnable parameters are the entries of $W$.
Let $W_*$ denote a minimizer of $L_{n,d}$, and define the following $d \times d$ matrices
\begin{eqnarray}
K:=n^{-1}X^\top X,\quad B:=n^{-1}X^\top Y.
\label{eq:KB}
\end{eqnarray}
When the matrix $K$ is invertible (which happens for large $n$), the normal equations give $W_*=K^{-1}B$.

For a fitted matrix $W$, the corresponding downstream predictor is $\widehat f(i)
    := \arg\max_{j\in[d]}(W^\top x)_j$,
where $x=\delta_i \in \mathbb R^d$ is the one-hot encoding of $i$.
Predictive performance can also be assessed using top-$k$ accuracy,
the proportion of examples for which the true next symbol appears
among the $k$ highest-scoring predictions, with $k=3$, for example.

% ...



\textbf{Scaling Laws for Gradient Descent.}
Consider a sequence of iterates $W(t)$ generated by gradient descent
(GD) initialized at zero, $W(0)=0$. For invertible $K$ in \eqref{eq:KB}, the regret
at initialization satisfies
\begin{align*}
    L_{n,d}(0)-L_{n,d}(W_*)
    &= \operatorname{tr}(W_*^\top K W_*)
     = \operatorname{tr}(B^\top K^{-1}B) \simeq \sum_{i=1}^d p_i
        \sum_{j=1}^d \pi_{j\mid i}^2
     = A,
\end{align*}
where $A\in(0,1]$ is the constant defined in \eqref{eq:A}.
The approximation replaces empirical frequencies by their population
counterparts in the large-sample regime.

\citet{kunstner2025scaling} studied as a function of optimization horizon $T$, the normalized regret measure
\[
    r_{n,d}(T)
    :=
    \frac{L_{n,d}(W(T))-L_{n,d}(W_*)}
         {L_{n,d}(0)-L_{n,d}(W_*)}.
\]
and distinguished three regimes of the Zipf exponent in
\eqref{eq:Zipf}: subcritical ($\alpha<1$), critical ($\alpha=1$),
and supercritical ($\alpha>1$). For each regime, they identified
appropriate scalings of the iteration count $T$ with vocabulary
size $d$ and derived explicit asymptotic laws for GD.
In particular, in the supercritical regime, their large-vocabulary analysis yields the epoch-wise scaling law
\[
    r_{n,d}(T)\asymp T^{-c},
    \qquad c:=1-1/\alpha,
\]
for large $T$. This exponent coincides with that obtained by
\citet{hutter2021learning,cabannes2024associative} for sample-size scaling, although the
resource varied here is the number of optimization steps.
The same work also derives scaling laws for sign gradient descent
(sign-GD), showing how the choice of optimizer changes the
dependence on vocabulary size.

These results describe optimization-time scaling under idealized
Zipfian frequency assumptions. Normalizing the regret factors out
the initialization scale $A$, which can itself depend on $d$,
while the limiting curves express the remaining vocabulary-size
dependence through a rescaling of time. Finite-sample fluctuations
are not tracked explicitly in this formulation.
Our results instead retain explicit dependence on vocabulary size
$d$ and sample size $n$, together with an independently adjustable
embedding width $N$, to characterize regret under a representational
constraint.

\subsection{A Capacity-Constrained Biagram Model}
\label{sec:RF-bigram}
In the model \eqref{eq:L} studied in \citep{kunstner2025scaling}. the number of trainable parameters equals $d^2$, with $d$ denoting the vocabulary size. However, this is unrealistic for the large-vocabulary setting considered here, which is also typical of real language models. To better reflect practice, we instead introduce a notion of capacity (i.e., parameter count) as follows. Given an embedding matrix $\Omega\in \mathbb R^{d \times N}$ with embedding dimension $N \le 1$, we assume the weights matrix $W$ admits the following factorization:\begin{eqnarray}
    W = \Omega\Theta,
\end{eqnarray}
with learnable factor $\Theta \in \mathbb R^{N \times d}$. This is a capacity constraint, and reduced the number of learnable parameters from $d^2$ to  $Nd \le d^2$.
Consider the modified optimization problem
\begin{equation}
    \operatorname{minimize } L_{n,d}(\Omega\Theta)\text{ over }\Theta \in \mathbb R^{N \times d},
    \label{eq:modified-L}
\end{equation}
where $L_{n,d}$ is the loss function defined in \eqref{eq:L}. 

\begin{assumption}[Random Embeddings]
The entries of the embedding matrix $\Omega$ are sampled iid from a distribution with zero mean, variance $1/d$, and with Gaussian concentration property. For example, the Gaussian distribution itself $\mathcal N(0,1/d)$.
\end{assumption}
Let $\hat \Theta_{n,d,N} \in \mathbb R^{N \times d}$ be any minimizer in \eqref{eq:modified-L} and set $\hat W_{n,d,N} := \Omega \hat \Theta_{n,d,N} \in \mathbb R^{d \times d}$. Our main focus will be the analysis of the f expected normalized regret:
\begin{equation}
\begin{split}
r_{n,d}(N) &:= \mathbb E[R_{n,d}(N)],\quad R_{n,d}(N) := \frac{L_{n,d}(\hat W_{n,d,N})-L_{n,d}(W_*)}{L_{n,d}(0) - L_{n,d}(W_*)},
        \end{split}
\label{eq:regret}
\end{equation}
where $A \in (0,1]$ is the constant in \eqref{eq:A}. We have made explicit the dependence on sample size $n$, vocabulary size $d$, and embedding dimension $N$, as these will appear explicitly in our scaling laws. 

\section{Main Results}
\label{sec:theory}

\subsection{Bias-Variance Decomposition}
Consider the vector $c=(c_1,\ldots,c_d) \in \mathbb R^d$ defined by
\begin{eqnarray}
c_k :=  \#\{t \in [n] \mid x_t = e_k\},
\end{eqnarray}
the number of times the $k$th vocabulary symbol appears in the dataset. Let $S$ be the number of unique vocabulary symbols present in the training dataset,
 \begin{eqnarray}
    S := \# I,\text{ where }I := \{k \in [d] \mid c_i>0 \}.
    \label{eq:S}
 \end{eqnarray}
Note that but $S$ and $I$ are deterministic functions of the vector of counts $c$. It is clear that $\sum_k c_k = n$ and $S \le d$. Finally, for any $k \in [d]$, let $\hat p_k := c_k/n$ be the empirical frequency of the $k$th token.


We start with the following general result.
\begin{proposition}
\label{prop:main}
Under Assumption~\ref{ass:main}, the conditional expected regret admits the following bias-variance decomposition:
\begin{equation}
    \boxed{
    \begin{split}
            &\mathbb E[R_{n,d}(N)\mid c]
    \simeq V+B,\quad B:=AN\nu,\quad V:=(1-A)\frac{(S-N)_+}{n},
    \end{split}
    }
    \label{eq:regret-decomp}
\end{equation}
where $A\in(0,1)$ is defined in \eqref{eq:A},
$\hat p_k:=c_k/n$, and $\nu=0$ when $N\ge S$.
For $0<N<S$, $\nu$ is the unique positive solution to the fixed-point equation
 \begin{eqnarray}
        N=\sum_{k\in I}\dfrac{\hat p_k}{\hat p_k+\nu}.
\end{eqnarray}
\end{proposition}

The second term, $V:=(1-A)(S-N)_+/n$ in the decomposition \eqref{eq:regret-decomp}
captures the finite-sample variability of the empirical transition
probabilities and can be interpreted as a \emph{variance contribution}.
It is present only when the embedding width $N$ is smaller than the
number $S$ of observed contexts, and vanishes once $N\ge S$.
Here, $S=S(c)\le\min(d,n)$ is the realized vocabulary occupancy
defined in \eqref{eq:S}; its distribution depends on the sample size
and the underlying context probabilities over the vocabulary.
In particular, the unconstrained bigram model has $N=d\ge S$,
and so $V=0$.

In contrast, the first term, $B:=AN\nu$
captures the \emph{approximation cost} of the representational bottleneck and can be interpreted as a bias contribution. This is reminiscent of model \emph{bias}. For $0<N<S$, the fixed-point equation gives the equivalent expression
$
    B=A\sum_{k\in I}
        \dfrac{\hat p_k \nu}{\hat p_k+\nu}.
$
Thus, unlike $V$, this term depends on the full empirical
context-frequency profile, rather than only on the number of
observed contexts. For a fixed frequency profile, it decreases as
the embedding width increases. However, it need not vanish as
the sample size grows when $N<d$: additional observations cannot
by themselves remove the limitations of a fixed representation.

Finally, when $N\ge S$, both terms in
\eqref{eq:regret-decomp} vanish. This agrees with the exact
interpolation property: whenever the embedding matrix restricted
to the observed contexts has full row rank, the student can
reproduce all empirical transition rows, yielding
$R_{n,d}(N)=0$.

\subsection{Sketch of Proof of Proposition \ref{prop:main}}
We provide the high-level ideas here. The complete proof is provided in the supplemental (appendix).

Conditioning on the empirical counts vector $c=(c_1,\ldots,c_d) \in \mathbb R^d$, one first show the exact identity
\begin{eqnarray}
    \mathbb E_{Y,\Omega}[R_{n,d}(N) \mid c] = \frac1n \sum_{k \in I} c_k\|M_k\|_2^2\,(1-h_k),
    \label{eq:step1}
\end{eqnarray}
where $ M_k$ is the $k$th row of the empirical matrix $M \in \mathbb R^{d \times d}$ of transition frequencies, i.e.
\begin{eqnarray*}
    M_{kk'} 
    := \frac{1}{c_k}\sum_{i=1}^n x_{ik} \, y_{ik'},
\end{eqnarray*}
and $h_k := \mathbb E[P_{kk}]$ is the expectation of the $k$th diagonal of entry of the matrix $P \in \mathbb R^{d \times d}$ which is the orthogonal projection onto the column space of $D_c^{1/2}\Omega$, with $D_c := \operatorname{Diag}(c)$.  One then computes explicitly
$$
c_k\|M_k\|^2 = c_k a_k + 1-a_k,\text{ with }a_k := \sum_{k'=1}^d \hat \pi_{k' \mid k}^2.
$$
By entry-wise concentration of $M$ to the next-word transition matrix $\pi$ (for example, see \citep{BerendKontorovich2012EmpiricalDistribution}), we know that $a_k \simeq \sum_{k'=1}^d \pi_{k' \mid k}^2 = A$ uniformly on $k \in [d]$, with high probability.
where $A \in (0,1]$ is the constant  appearing in equation \eqref{eq:A} of Assumption \ref{ass:main}. It follows from \eqref{eq:step1} that 
$$
\mathbb E[R_{n,d}(N) \mid c] \simeq   \frac{A}{n} \sum_{k \in I} (1-h_k)c_k + \frac{1-A}{n} \sum_{k \in I} (1-h_k).
$$
Finally, using random matrix theory, one shows that $h_k \simeq c_k/(c_k + N\nu)$ uniformly on $k$, where $\nu > 0$ is as defined in the proposition. Plugging this estimate into the above equation and rearranging then gives the desired result. \qed

\subsection{Explicit Scaling Laws}
Note that the RHS in \eqref{eq:regret-decomp} is random as it depends on the random counts $c=(c_1,\ldots,c_d)$. In order to de-randomize the formula, we need the sample size $n$ to be sufficiently large as a function of the vocabulary size $d$. Let $\alpha$  the Zipfian exponent for the marginals as in Assumption \ref{ass:main}.

\begin{theorem}
Consider the sublinear regime where the random embedding dimension $N \le d$ grows with the vocabulary size $d$. Finally, assume sample size $n$ is sufficiently large in the sense that $n \gg d^\alpha H_{d,\alpha}$. Then, we have the followin exact scaling law for the regret:
\begin{eqnarray}
\boxed{
\begin{split}
       r_{n,d}(N) &\simeq B + V,\text{ with }B := A N\nu,\text{ and }
        V := 
        (1-A)\begin{cases}
        \phi,&\mbox{ if }N = o(d),\\
        (1-\psi)\phi,&\mbox{ if }N \asymp d,
    \end{cases}
\end{split}
}
\label{eq:exact}
\end{eqnarray}
where the constant $A \in (0, 1]$ is the statistic defined in \eqref{eq:A}, while $\psi = \psi(N,d) := N/d$, $\phi=\phi(n,d):=d/n$, and
$\nu = \nu_\alpha(N,d)$ is the unique positive solution to the following fixed-point equation
\begin{eqnarray}
    N = \sum_{i=1}^d \dfrac{p_i}{p_i + \nu}.
 \end{eqnarray}

Furthermore, we have the following breakdown:
\begin{itemize}
    \item [(i)] \textbf{(Sub-critical Zipf exponent).} If the exponent in \eqref{eq:Zipf} verifies $0 \le \alpha < 1$, then
    \begin{eqnarray}
   B  \simeq  A\begin{cases}
                 1,&\mbox{ if }N = o(d),\\
            (1-\alpha)\psi/\rho,&\mbox{ if }N \asymp d
        \end{cases}
    \end{eqnarray}
where  $\rho=\rho(\alpha,\psi)$ is the unique positive solution to the fixed-point equation 
\begin{eqnarray}
    \psi = \int_0^1 \dfrac{\rho}{\rho + t^\alpha}\mathrm d t.
\end{eqnarray}
\item[(ii)] \textbf{(Critical Zipf exponent).} If $\alpha=1$, then
    \begin{eqnarray}
       B \simeq \frac{A}{\log d}\begin{cases}
            \log(d/N),&\mbox{ if }N=o(d),\\
            1/q,&\mbox{ if }N \asymp d,
        \end{cases}
        \label{eq:alpha1}
    \end{eqnarray}
where
is the unique positive solution to the following fixed-point equation:
\begin{eqnarray}
    N = \sum_{i=1}^d \dfrac{q}{q+i}.
\end{eqnarray}

\item[(iii)] \textbf{(Super-critical Zipf exponent).} If $\alpha > 1$, then $B \simeq  C_\alpha N^{-(\alpha-1)}$,
where $C_\alpha$ is a positive constant only depending on $\alpha$.
\end{itemize}
\label{thm:finite-n}
\end{theorem}
In the super-critical regime ($\alpha>1$), the $\alpha-1$ exponent in our  scaling law $B \simeq N^{-(\alpha-1)}$ the one reported in one of the main results in \citep{cabannes2024associative}. In the critical regime ($\alpha=1$), let $b=-\mathcal W_{-1}(-\psi e^{-\psi})-\psi$, where $\mathcal W_{-1}$ denotes the $-1$ branch of the Lambert--W function. Then the $1/q$ term in \eqref{eq:alpha1} admits the following expansion:
$$
\dfrac{1}{q} \simeq b - \dfrac{b^2}{b-1}\,\dfrac{1}{2N}.
$$

Note that the condition $n \gg d^\alpha H_{d,\alpha}$ is very mild and in satisfied in language datasets like OpenWebText2 \citep{openwebtext}, WikiText-2 \citep{wikitext}, where $\alpha < 1$ (refer to Table \ref{tab:params}), and so $H_{d,\alpha} = O(\log d)$; the condition then simply requires the sample size $n$ to grow super-linearly with vocabulary size $d$. This condition ensures that the dataset contains enough information to allow for the estimation of the head of the marginal distribution of the tokens in the vocabulary.

In the sub-critical and critical regimes which better capture to natural language distribution, our theorem provides new scaling laws which better match the empirically observed laws of  \citet{kaplan2020scaling}. 

\subsection{A Canonical Scaling for the Embedding Dimension}
A reasonable embedding dimension $N$ w.r.t the vocabulary size $d$ is
\begin{eqnarray}
    N \simeq \gamma d^\beta,
    \label{eq:Nscaling}
\end{eqnarray}
for constants $\gamma,\beta>0$ with $\beta \le 1$. In particular, $\beta=1$ captures the linear scaling regime where $N \asymp d$, while $\beta \in (0,1)$ captures the sublinear regime where $N=o(d)$. By fixing $d$ and sample size $n$, we empirically trace the decay of the excess risk and overlay the exact closed-form asymptotics derived from our random matrix theory approach. For example, in the critical setting where the marginal Zipf exponent is $\alpha=1$, the above theorem predicts that provided the sample size is sufficiently large (i.e. $n \gg d\log d$), the quenched regret is given by 
\begin{eqnarray}
    r_{n,d}(N) \simeq \begin{cases}
        A(1-\beta) + (1-A)\phi,&\mbox{ if }\beta < 1,\\
        A/(q\log d) + (1-A)(1-\gamma)\phi,&\mbox{ if }\beta=1,
    \end{cases}
\end{eqnarray}
with $\phi := d/n$.

\input{experiments}
\input{conclusion}

\subsection*{Acknowledgement} We would like to acknowledge that the first author \textbf{Julian Enoch} is a PhD student funded by bursaries from Canada Research Chair grant number CRC-2023-00228 (\emph{Computing for the Greater Good}) awarded to \textbf{Elvis Dohmatob}.

\subsection*{AI use statement}
In this work, we have made limited use of generative AI tools for spell-checking, and also reformulating chunks of text in the introduction (Section \ref{sec:intro}, specially \ref{sec:related-work}). Also, AI was used to help with the tedious verification / sanity-checking of some of our proofs (Section \ref{sec:proofs}). No other use of AI was made in our work. The ideas and execution involved in this work is ours. We take full responsibility for the final content of this work, including text, claims, or artifacts produced with the aid of AI.

\bibliography{refs.bib}

\clearpage

\appendix

\clearpage
\thispagestyle{empty}
\vspace*{\fill}

\begin{center}
    {\Huge\bfseries Appendix}
\end{center}

\input{proofs}

\end{document}

%% file: intro.tex
\section{Introduction}
\label{sec:intro}
Neural scaling laws describe systematic relationships between predictive
performance, model size, training data, and compute, providing an empirical
basis for allocating resources in language-model training
\citep{kaplan2020scaling}. Theoretical studies have connected these
relationships to data distributions, model capacity, and optimization
dynamics \citep{maloney2022solvable,cabannes2024associative,kunstner2025scaling}.
An important challenge is to distinguish these mechanisms: similar
learning curves can reflect insufficient observations, a restricted
representation, or incomplete optimization.

In language modeling, the highly uneven frequencies of tokens make the
role of representational capacity particularly interesting. An
unconstrained bigram model assigns a separate next-token distribution
to every context, regardless of how frequently that context occurs.
Does accurate prediction require this degree of flexibility, or can
a substantially smaller representation suffice? More broadly, how
does the power-law structure of language determine the cost of
compressing its predictive information? Addressing these questions
provides a way to connect empirical capacity scaling to explicit
properties of the underlying data distribution.

\subsection{Contributions}

Our main contributions can be summarized as follows.

\textbf{-- A bottlenecked bigram model.}
We introduce a capacity-constrained bigram model with Zipfian token statistics
as an analytically tractable toy model of a large language model.
Fixed random embeddings serve as a proxy for representational capacity,
imposing a width-$N$ bottleneck with $dN$ trainable parameters.
By studying the optimally fitted predictor, we isolate the cost of
limited capacity from optimization error. This extends the unrestricted
bigram setting of \citet{kunstner2025scaling} by introducing an explicit
representational constraint, complementing their analysis of scaling
with the number of optimization steps. In the summable Zipf regime
$\alpha>1$, our results also recover the associative-memory scaling
laws of \citet{cabannes2024associative}.

\textbf{-- Exact scaling laws.}
    We obtain an exact representation of regret relative to the
    unconstrained empirical optimum and derive asymptotic scaling laws
    in vocabulary size $d$, sample size $n$, embedding dimension $N$,
    and the Zipf exponent. The parameterization $N=\gamma d^\beta$
    reveals distinct sublinear and linear capacity regimes. In the
    appropriate regime, the resulting laws recover the functional
    form of classical empirical neural scaling laws
    \citep{kaplan2020scaling}.

 \textbf{-- An explicit account of finite-sample effects.}
    Our theory incorporates vocabulary occupancy and variability in
    empirical transition probabilities, separating a capacity-dependent
    contribution to regret from a finite-sample correction. This identifies when capacity constraints dominate and when limited observations materially affect the scaling behavior.

\textbf{-- Empirical evaluation.}
    In Section \ref{sec:exp}, we present results of extensive experiments which validate our scaling laws both on 
    real-world natural language datasets, WikiText-2 \citep{wikitext} and OpenWebText2 \citep{openwebtext}, support
    the predicted scaling curves across vocabulary sizes $d$, sample sizes $n$, and embedding dimensions $N$. They examine both the decay of regret under sublinear capacity and its approach to zero as the bottleneck becomes sufficiently wide.

    \input{related}

%% file: related.tex
\subsection{Related work}
\label{sec:related-work}
The pioneering work of \citet{kaplan2020scaling} established empirical relationships between language-model loss, parameter count, training data, and compute, motivating the use of scaling laws for resource allocation. Subsequent theoretical work has identified several mechanisms that can produce such behavior. 

\textbf{Zipfian statistics, sampling, and optimization.}
\citet{hutter2021learning} derive data-dependent learning curves for a memorization-based model, explaining how power-law frequencies lead to power-law decay of prediction error with sample size. More recently, \citet{kunstner2025scaling} analyze gradient descent and sign descent for a linear bigram model under Zipfian assumptions, characterizing how optimization depends on vocabulary size and the frequency exponent. Their results emphasize the importance of the regime near $\alpha=1$. Neither analysis introduces a representation width that can be varied independently of vocabulary size: Hutter focuses on sample-size scaling, while Kunstner and Bach study optimization of an unrestricted bigram parameterization. Consequently, they do not characterize the approximation error caused by a separate embedding-width budget. We introduce that capacity axis through a fixed random embedding of width $N$, distinguishing representational limitations from finite-data and optimization effects.

\textbf{Finite-capacity associative memories.}
\citet{cabannes2024associative} provide a closely related capacity-based perspective, deriving sample- and parameter-scaling laws for associative memories built from outer products of random embeddings. Their analysis explicitly addresses finite memory and the efficiency of different storage schemes. The principal infinite-vocabulary power-law results concern the summable regime $\alpha>1$. In that regime, the Zipf normalizing constant has a finite limit as vocabulary size grows, allowing leading-order laws that do not retain an explicit vocabulary cutoff. This contrasts with $\alpha\leq1$, where normalization remains vocabulary-dependent. Our analysis keeps vocabulary size $d$ explicit and studies its interaction with the bottleneck width, including sublinear and linear parameterizations $N=\gamma d^\beta$. We also study squared-loss regret for conditional next-token probabilities rather than the association-retrieval error analyzed in that work.

\textbf{Spectral and random-feature theories.}
Kernel and random-feature models provide another route to analytical scaling laws. \citet{cui2021generalization} characterize how spectral decay, target regularity, noise, and regularization determine learning-curve exponents, including the crossover between noiseless and noisy regimes. \citet{maloney2022solvable} develop a solvable generative and random-feature model with joint dependence on dataset size and parameter count, and explain how a finite spectral range can interrupt power-law scaling. These works already incorporate statistical capacity constraints; our contribution is not to introduce capacity into scaling-law theory in general. Rather, we connect an explicit random-embedding bottleneck to discrete Zipfian token statistics and characterize regret relative to the unconstrained empirical bigram optimum. This distinction also matters when comparing our results with population generalization error or cross-entropy scaling in other models.

\citet{bahri2024explaining} distinguish variance-limited and resolution-limited regimes and relate scaling exponents to spectral properties of the learning problem. These results indicate that superficially similar power laws can arise from different statistical or computational limitations. Our work studies one such mechanism in a discrete next-token prediction setting: the error induced by a restricted representation, after optimizing the predictor within that representation.

\textbf{Theory of skill acquisition in language models.}
\citet{michaud2023quantization} propose that smooth aggregate scaling can emerge from learning discrete skills with a heavy-tailed frequency distribution. \citet{nam2024exactly} develop an exactly solvable model for multitask sparse parity, connecting skill emergence with scaling in training time, data, and model size. In a stochastic-process formulation, \citet{debowski2023simplistic} construct multiperiodic Santa Fe processes and memory-based predictors exhibiting power-law cross-entropy decay with model and data resources. These approaches explain scaling through the acquisition of discrete information. Our model instead studies the approximation of empirical next-token distributions through a shared, finite-dimensional random representation, without postulating a sequence of separately acquired skills.

\textbf{Distributional tails and model collapse.}
\citet{dohmatob2024tale} study how incorporating synthetic data into
training distributions changes scaling laws and can lead to model
collapse \citep{shumailov2024aimodelscollapse}. Their analysis highlights the importance of preserving
distributional tails: increasing data or model size need not yield
the same improvements when rare events are suppressed.
This provides a complementary distributional perspective on scaling,
whereas our work studies the regret induced by a representational
bottleneck under a specified Zipfian distribution.

\textbf{Attention models and $n$-gram structure.}
\citet{lyu2025attention} derive scaling laws for a solvable self-attention
model, characterizing dependencies on training time, model size, data,
and context length. From a complementary empirical perspective,
\citet{nguyen2024ngrams} investigate how well rules based on $n$-gram
statistics approximate transformer next-token predictions.
These studies motivate simplified sequence-prediction models as
analytical testbeds, without equating them with full transformers.
Our bottlenecked bigram model isolates the interaction between
Zipfian token statistics and finite representational capacity,
rather than modeling attention or long-context dependencies.

%% file: experiments.tex
\section{Empirical Validation}
\label{sec:exp}
\subsection{Experimental Setup}
We validate the capacity--risk predictions from Section~\ref{sec:theory} in a setting where representational capacity can be tuned independently of corpus statistics. We study the capacity-constrained linear bigram model in which the target embeddings are restricted to a rank-$N$ subspace via a fixed random projection. For each configuration, we optimize to convergence and report the normalized excess risk (regret) $r_{n,d}(N)$ relative to the unconstrained empirical optimum (cf.~\eqref{eq:exact}).

To connect with the broader scaling-laws literature \citep{kaplan2020scaling}, we treat $N$ as an explicit capacity axis and couple it to vocabulary size through $N=\gamma d^{\beta}$ (\eqref{eq:Nscaling}), sweeping $\gamma$ and $\beta$ to probe both the linear-width regime ($\beta=1$) and sublinear-width regime ($\beta<1$).

\begin{figure*}[!htbp]
    \centering
    \begin{subfigure}{1\textwidth}
        \centering
        \includegraphics[width=.45\linewidth]{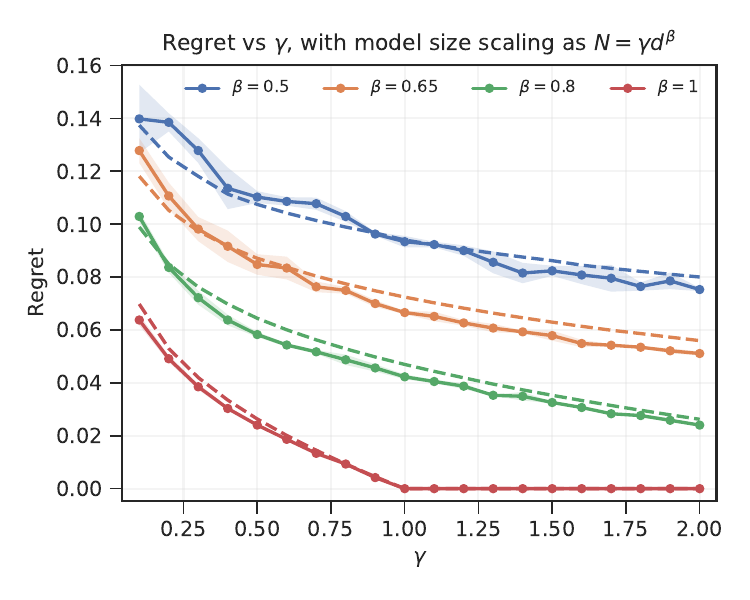}
        \includegraphics[width=.45\linewidth]{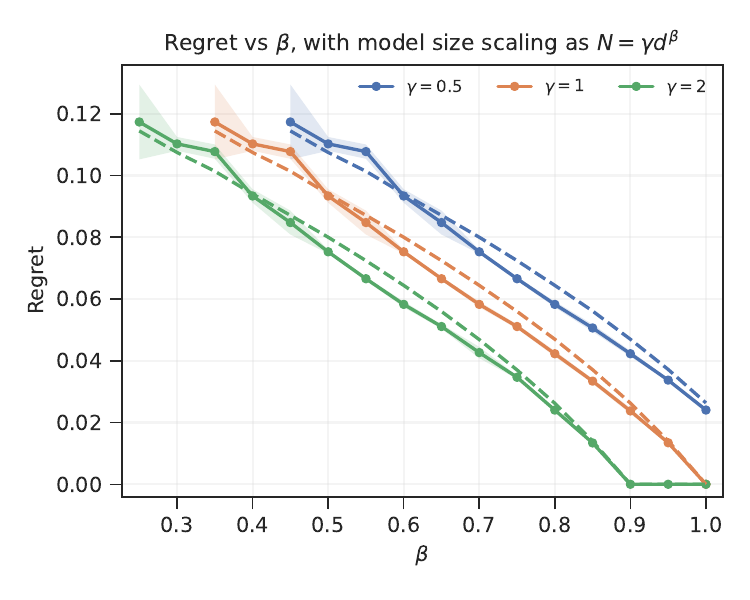}
        \vspace{-.3cm}
\caption{Vocabulary size $d=10^3$ and sample size $n=10^5$.}
        \label{fig:wikitext-d1000}
    \end{subfigure}

    \par\medskip

    \begin{subfigure}{1\textwidth}
        \centering
        \includegraphics[width=.45\linewidth]{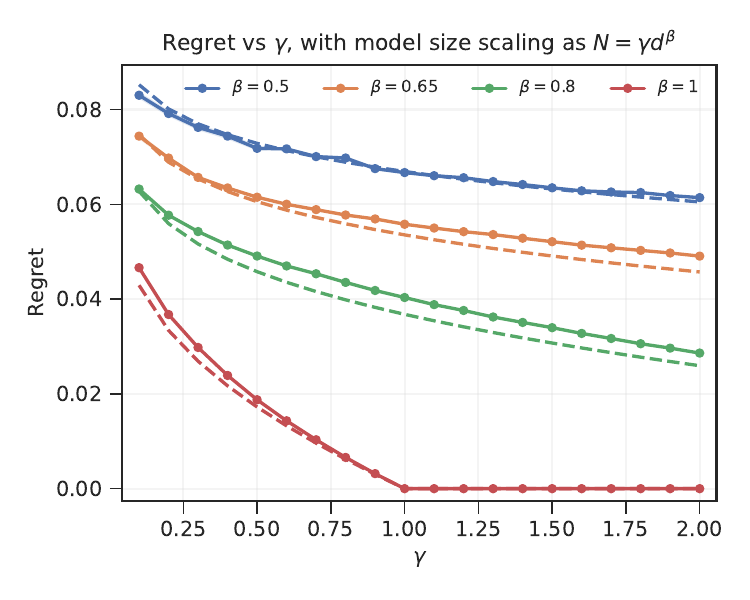}
        \includegraphics[width=.45\linewidth]{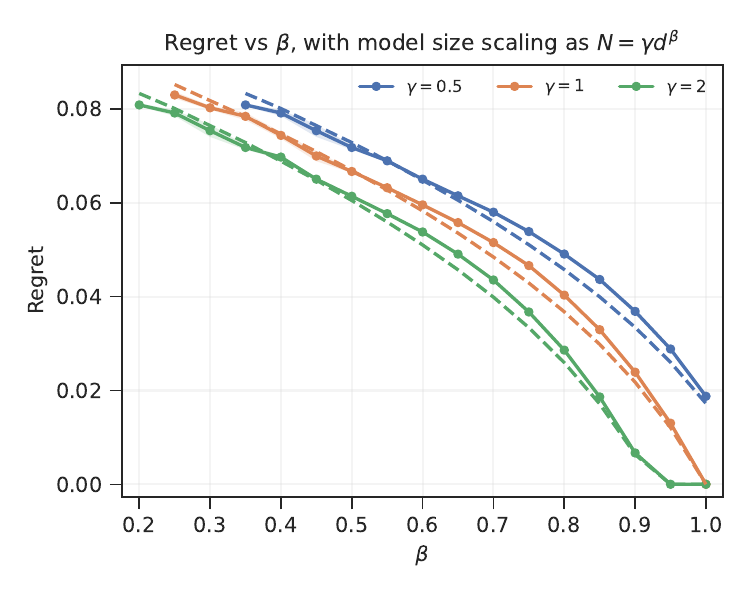}
            \vspace{-.3cm}
        \caption{Vocabulary size $d=10^4$ and sample size $n=10^6$.}
        \label{fig:wikitext-d10000}
    \end{subfigure}
\caption{\textbf{Capacity scaling on WikiText-2.} Converged normalized excess risk (regret) $r_{n,d}(N)$ \eqref{eq:regret} for the random-features bigram model, as the embedding width is scaled as $N=\gamma d^\beta$ (\eqref{eq:Nscaling}). \textbf{Left:} vary the width multiplier $\gamma$ at fixed exponents $\beta$. \textbf{Right:} vary the exponent $\beta$ at fixed multipliers $\gamma$. Solid curves (markers) show empirical means over random projections; shaded regions indicate variability across trials. Dashed curves are the theoretical predictions from our Theorem~\ref{thm:finite-n}. Notice the excellent match.}
\end{figure*}

\begin{figure*}[!htbp]
    \centering
    \begin{subfigure}{\textwidth}
        \centering
        \includegraphics[width=.45\linewidth]{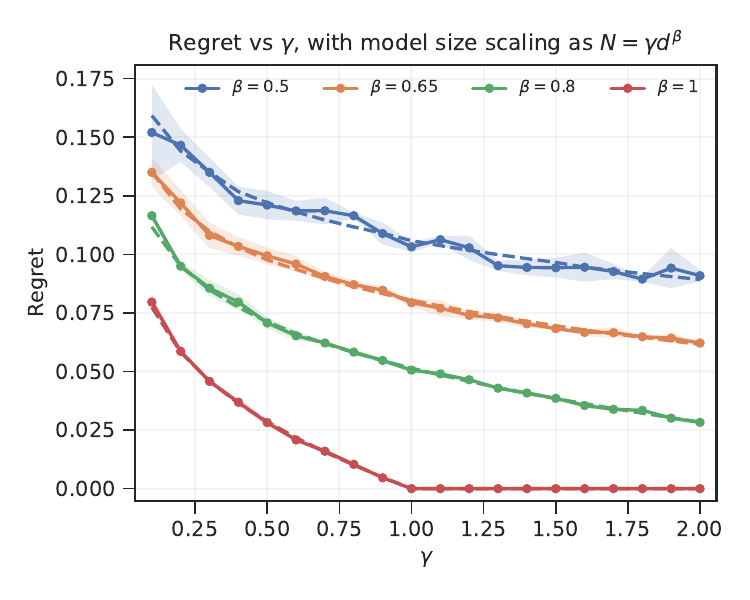}
        \includegraphics[width=.45\linewidth]{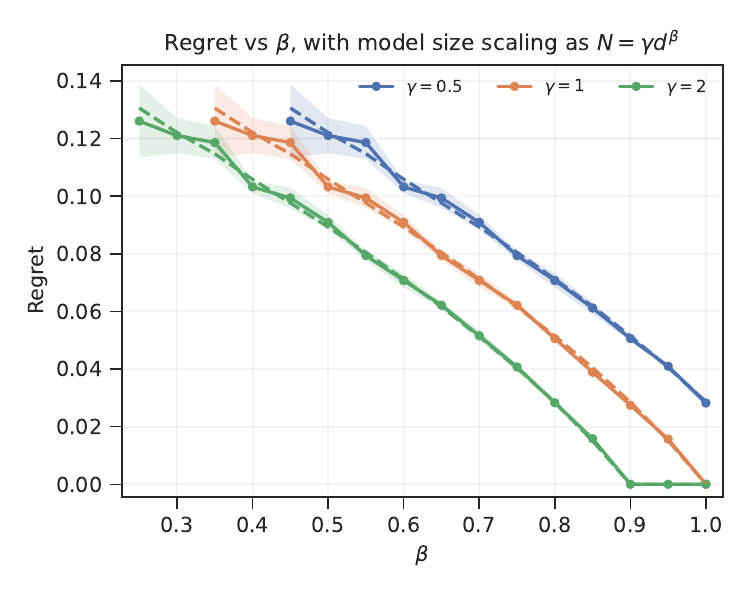}
        \vspace{-.3cm}
        \caption{Vocabulary size $d=10^3$ and sample size $n=10^5$.}
        \label{fig:openwebtext-d1000}
    \end{subfigure}

    \par\medskip

    \begin{subfigure}{\textwidth}
        \centering
        \includegraphics[width=.45\linewidth]{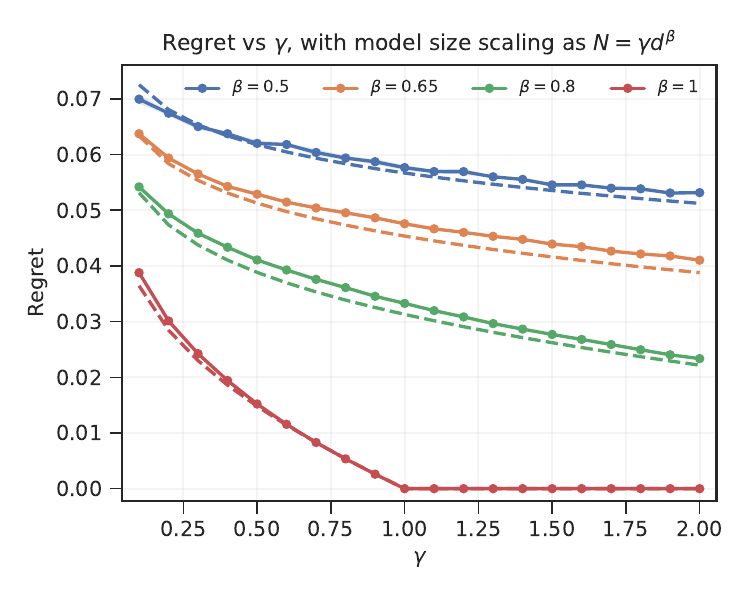}
        \includegraphics[width=.45\linewidth]{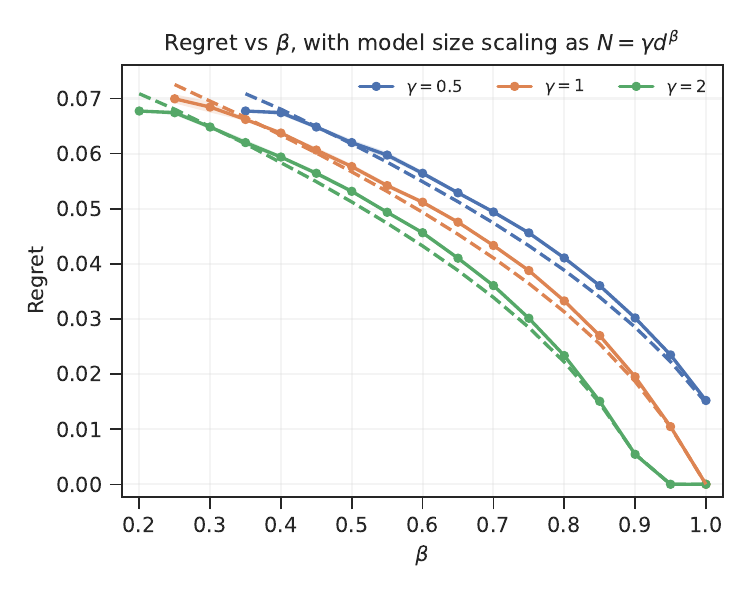}
            \vspace{-.3cm}
        \caption{Vocabulary size $d=10^4$ and sample size $n=10^6$.}
        \label{fig:openwebtext-d10000}
    \end{subfigure}

    \caption{\textbf{Capacity scaling on OpenWebText2.} Converged normalized excess risk (regret) $r_{n,d}(N)$ \eqref{eq:regret} for the random-features bigram model with a rank-$N$ embedding bottleneck, where $N=\gamma d^\beta$ (\eqref{eq:Nscaling}). \textbf{Left:} vary $\gamma$ at fixed $\beta$. \textbf{Right:} vary $\beta$ at fixed $\gamma$. Solid curves (markers) show empirical means over random projections; shaded regions show trial variability. Dashed curves are the corresponding theoretical predictions from our Theorem~\ref{thm:finite-n}. Notice the excellent match.}
    \label{fig:openwebtext-capacity-scaling}
\end{figure*}

\subsection{Datasets}
We run experiments across three distinct datasets: Synthetic data generated exactly from Zipfian distributions, and the empirical real-world corpora WikiText-2 \citep{wikitext} and OpenWebText2 \citep{openwebtext}. For the real-world datasets, the vocabulary size $d$ and the token count $n$ are calculated over the whole dataset. 
The Zipfian exponent $\alpha$ are extracted through log-log regression over the corpus distribution. Table \ref{tab:params} provides a list of these descriptive statistics.

\subsection{Results}
The results across all configurations showcase an alignment between the empirical excess risk of the capacity-constrained models and our theoretical predictions. Despite the noise inherent in real-world distributions, the derived asymptotics capture the empirical behavior across all datasets.



\textbf{WikiText-2.}
Figure \ref{fig:wikitext-d1000} and \ref{fig:wikitext-d10000} demonstrate the capacity scaling laws for the WikiText-2 dataset \citep{wikitext}, for different values of the vocabulary size $d$ and embedding dimension $N$, for different values of $\gamma$ and $\beta$ in \eqref{eq:Nscaling}. Notice how excellently our theory predicts the performance of the model. From the figures, we see that increasing capacity $N$ increases the model capacity and reduces regret in both settings. In the linear scaling regime ($\beta=1$), the model reaches zero regret at $\gamma=1$, while in the sublinear scaling regime ($\beta < 1$)  it retain nonzero regret floor. 

\paragraph{OpenWebText2.}
Figure~\ref{fig:openwebtext-capacity-scaling} shows the same capacity scaling analysis on OpenWebText2 \citep{openwebtext} for two vocabulary cutoffs. Despite the substantially noisier empirical frequency profile of a web-scale corpus (and the resulting deviations from an ideal Zipf law at the head and tail), the predicted capacity--risk tradeoff remains remarkably accurate.

Qualitatively, we again observe the two regimes predicted by Theorem~\ref{thm:finite-n}. First, in the linear-width regime ($\beta=1$), increasing the width multiplier $\gamma$ produces a sharp decrease in regret, approaching the unconstrained optimum once $N$ is on the order of $d$. Second, in the sublinear regime ($\beta<1$), improvements with $\gamma$ diminish and the curves flatten, indicating that performance becomes variance-limited rather than approximation-limited. This saturation is consistent with the finite-sample term in \eqref{eq:exact}, which scales with $d/n$ and therefore persists even as $N$ increases when $N$ grows too slowly with $d$.

Beyond confirming the asymptotics on cleaner benchmarks (WikiText-2) and synthetic data, OpenWebText2 provides evidence that the scaling-law picture is robust to realistic corpus artifacts. In particular, the close match between empirical curves and the theorem-derived predictions suggests that the dominant driver of the observed scaling is the interaction between heavy-tailed token frequencies and the explicit embedding bottleneck, rather than details of text domain or preprocessing.

%% file: conclusion.tex
\section{Concluding Remark}
Our results isolate representational capacity as a source of scaling,
complementing explanations based on optimization dynamics. Keeping
vocabulary size explicit distinguishes increasing the capacity available
for a fixed prediction problem from enlarging the problem itself.
This distinction is particularly relevant near the critical Zipf
exponent $\alpha=1$, where vocabulary-dependent normalization affects
the scaling behavior. Together, these perspectives compare empirically supported believes that
heavy-tailed language statistics may underlie both capacity- and optimization-driven scaling \citep{kaplan2020scaling}.

Our analytically tractable setting provides a quantitative benchmark for capacity-driven scaling, while experiments on synthetic and natural language data support the predicted dependence on model width.
This framework opens the way to studying how learned embeddings, longer contexts, and finite optimization budgets modify these laws. Such extensions would help identify which scaling behaviors reflect the underlying language statistics and which depend on architectural
and training choices.

\paragraph{Future Directions.}
A natural extension is to replace fixed random embeddings with learned representations and investigate how adaptation to predictive structure changes the capacity scaling.

Also, extending the analysis to higher-order token dependencies and heterogeneous transition distributions would further reveal how context length and richer language statistics shape these tradeoffs.

%% file: proofs.tex
\section{Proof of Main Results}
\label{sec:proofs}
\subsection{Proof of Proposition \ref{prop:main}}
As in the proposition, define $c_k := \#\{t \in [n] \mid x_t = e_k\}$, $I = \{ k \in [d] \mid c_k > 0\}$, and $S:=|I|$, where $x_t$ is the $t$th row of the design matrix $X \in \mathbb R^{n \times d}$. For any candidate student model $W \in \mathbb R^{d \times d}$, one can write the loss as
\begin{eqnarray}
    L(W) := \frac{1}{n}\|XW - Y\|_F^2 = \frac{1}{n}\sum_{k \in I} \sum_{x_t = e_k}\|W_k - y_t\|_2^2.
\end{eqnarray}

\paragraph{Step 1: A Formula for Regret of any Student.} Define an $S \times d$ matrix $M$ by setting
\begin{eqnarray}
M_k := \frac{1}{c_k}\sum_{x_t = e_k}y_t,\text{ for all }k \in I.
\end{eqnarray}
The $k'$th entry of $M_k$ is nothing but the (empirical) probability of observing the $k'$th token immediately after the $k$th token. We can write $W_k - y_t = (W_k - M_k) + (M_k - y_t)$, and so
$$
\sum_{x_t=e_k}\|W_k - y_t\|_2^2 = c_k\|W_k - M_k\|_2^2 + \sum_{x_t=e_k}\|M_k - y_t\|_2^2 + 2(W_k-M_k)^\top \sum_{x_t=e_k}(M_k - y_t).
$$
But, $\sum_{x_t=e_k}(M_k - y_t) = 0$ by definition of $M_k$. We deduce that
\begin{eqnarray*}
    L(W) = \frac{1}{n}\sum_{k \in I}c_k \|W_k - M_k\|_2^2 + \frac{1}{n}\sum_{k \in I}\sum_{x_t=e_k}\|M_k - y_t\|_2^2.
\end{eqnarray*}
Since an unconstrained minimizer $W_*$ satisfies $W_{*,k} = M_k$ for all $k \in I$, we deduce that the regret of any $W$ is given by the following exact formula:
\begin{eqnarray}
    R(W) := L(W) - L(W_*) = \frac{1}{n}\sum_{k \in I}c_k \|W_k - M_k\|_2^2.
\end{eqnarray}

\paragraph{Step 2: Representation of the Optimal Student and Its Regret.} We now restrict to students of the form $W = \Omega\Theta$ for some $\Theta \in \mathbb R^{N \times d}$. Let $\Omega_I \in \mathbb R^{S \times d}$ be the restriction of $\Omega$ to the rows corresponding to observed tokens $k \in I$. Define
$$
D_c := \operatorname{diag}(c_k \mid k \in I),\quad A_c := D_c^{1/2}\Omega_I,\quad B_c := D_c^{1/2} M.
$$
Then, the constrained estimator is $\hat W = \Omega\hat \Theta$, where $\hat \Theta$ minimizes
$$
\frac1n \sum_{k \in I}c_k \|(\Omega\Theta)_k - M_k\|_2^2 = \frac1n \|A_c\Theta - B_c\|_F^2.
$$
The normal equations for $\hat \Theta$ are $A_c^\top A_c\hat\Theta = A_c^\top B_c$. This gives
$$
A_c\hat\Theta = P_{\Omega_I} B_c,\text{ with }P_{\Omega_I} := A_c (A_c^\top A_c)^\dagger A_c^\top = D_c^{1/2}\Omega_I(\Omega_I^\top D_c \Omega_I)^\dagger \Omega_I^\top D_c^{1/2}.
$$
Note that $P_{\Omega_I}$ is the orthogonal projection matrix onto the column space of $A_c = D_c^{1/2}\Omega_I$. Defining $P_{\Omega_I}^\perp := \operatorname{I} - P_{\Omega_I}$, one has
$$
R(W) = \frac1n\|P_{\Omega_I}^\perp B_c\|_F^2 = \frac{1}{n}\trace B_c^\top P_{\Omega_I}^\perp B_c = \frac1n \trace (\operatorname{I}_S - P_{\Omega_I})B_cB_c^\top.
$$
We now marginalize of $\Omega$. We shall show that $\mathbb E_{Y, \Omega}[P_{\Omega_I}]$ is diagonal. To this end, define
$$
h_k := \mathbb E_{Y, \Omega}[(P_{\Omega_I})_{k,k}].
$$
These scalars should be reminiscent of \emph{leverage scores} in sketching methods.
Let $J=\operatorname{diag}(\varepsilon_1,\ldots,\varepsilon_S)$, with $\varepsilon_1,\ldots,\varepsilon_S \in \{-1,1\}$. Since $J$ commutes with $D_c$ (both being diagonal matrices), one has $JD_c J = J^2 D_c = D_c$ and so
$$
P_{J \Omega_I}  = D_c^{1/2} J\Omega_I (\Omega_I^\top J D_c J \Omega_I)^\dagger \Omega_I^\top J D_c^{1/2} = J D_c^{1/2} \Omega_I (\Omega_I^\top D_c \Omega_I)^\dagger \Omega_I^\top D_c^{1/2} J = J P_{\Omega_I} J.
$$
On the other hand, due to the rotational invariance of the distribution of $\Omega$, we know that $J\Omega_I = J$ in distribution, and so
$$
\mathbb E_{Y, \Omega}[P_{\Omega_I}] = \mathbb E_{Y, \Omega} [P_{J \Omega_I}] = J\, \mathbb E_{Y, \Omega} [P_{\Omega_I}] J.
$$
The above identity holds for any diagonal matrix with diagonal entries $\pm 1$. For any $k \ne k'$, choose $J$ so that it flips the the signs of the $k$th but not the $k'$th row of $P_{\Omega_I}$. Then, $(\mathbb E_{Y, \Omega}[P_{\Omega_I}])_{ij} = (J\, \mathbb E_{Y, \Omega} [P_{\Omega_I}] J)_{ij} = -(\mathbb E_{Y, \Omega}[P_{\Omega_I}])_{ij}$, and so $(\mathbb E_{Y, \Omega}[P_{\Omega_I}])_{ij} = 0$. We conclude that
\begin{eqnarray}
    \mathbb E_{\Omega} [P_{\Omega_I}] = \operatorname{diag}(h_k \mid k \in I).
\end{eqnarray}
This in turn gives
$$
\mathbb E_{Y, \Omega}[r_{n,d}(N)] = \frac1n \trace (\operatorname{I}_S - \mathbb E_{Y, \Omega}[P_{\Omega_I}])D_c^{1/2} M M^\top D_c^{1/2} = \frac1n \sum_{k \in I} c_k\|M_k\|_2^2\,(1-h_k).
$$
Now, one computes
\begin{align*}
    \|M_k\|^2 &= \frac{1}{c_k^2}(\sum_{t | x_t = e_k} \mathbb E[\|y_t\|^2] + \sum_{t' \ne t,\,x_{t'} = x_t = e_k}\mathbb E[y_{t'}^\top y_t]) = \frac{1}{c_k^2}(c_k + c_k(c_k-1)a_k)\\
    &= a_k + \frac{1-a_k}{c_k},
\end{align*}
where we have used the fact that $\|y_t\|^2 = 1$ for all $t$ because the $y_t$'s are on-hot encodings, and $\mathbb E[y_{t'}^\top y_t] = \mathbb E[y_{t'}]^\top \mathbb E[y_t] = a_k := \|\mathbb E[y \mid x = e_k]\|^2$ for $t' \ne t$ due to independence. under Assumption \ref{ass:main}, we have $a_k \equiv A$, and so
\begin{align*}
\mathbb E_{Y, \Omega}[r_{n,d}(N) \mid c] = \sum_{k \in I}\|M_k\|_2^2\,(1-h_k) = \frac A n \sum_{k \in I} c_k (1-h_k) + \frac{1-A}{n}\sum_{k \in I}1-h_k.
\end{align*}
Now, $\sum_{k \in I}h_k$ equals the rank of $P_{\Omega_I}$ which is $\min(S,N)$ a.s. We deduce that $\sum_{k \in I}(1-h_k) = S-\min(S,N) = (S-N)_+$, and so 
$$
\mathbb E_{Y, \Omega}[r_{n,d}(N)] =  \frac A n \mathbb E\sum_{k \in I} c_k \, (1-h_k) + (1-A)\mathbb E \frac{(S-N)_+}{n}
$$

\paragraph{Step 3: Random Matrix Theory.} We now produce a deterministic equivalent for the random quantity $\sum_{k \in I}c_k\,(1-h_k)$. Observe that if $N < S := |I|$, then the rank of $\Omega_I$ is $S$; $G := \Omega_I' \Omega_I$ has full rank $N$, and is therefore invertible a.s. Thus, for any $k \in I$, one has $h_k = c_k w_k' G^{-1}w_k$, where $w_k$ is row of $\Omega$ corresponding to the $k$th token.

We now use a leave-one-out (LOO) argument. Decompose the covariance matrix $G = c_k  w_k w_k^\top + G_{-k}$, where $G_{-k} := \sum_{\ell \ne k}c_\ell  w_\ell w_\ell^\top$. The Sherman-Morrison formula gives
\begin{align*}
    G^{-1} = G_{-k}^{-1} - \frac{c_k}{1+c_k  w_k^\top G_{-k}^{-1}  w_k}G_{-k}^{-1}w_k w_k^\top G_{-k}^{-1}.
\end{align*}
Plugging into the previous expression for $\ell_i$ and simplifying gives
\begin{eqnarray*}
    1-h_k = \frac{1}{1+c_k  w_k^\top G_{-k}^{-1} w_k} = \frac{1}{1+(c_k/k)\trace G_{-k}^{-1}} + o(1).
\end{eqnarray*}
The last step is because $w_k$ and $G_{-k}$ are independent, so that $w_k^\top G_{-k}^{-1} w_k = d^{-1}\trace G_{-k}^{-1} + o(1)$. But $(1/d)\trace G_{-k}^{-1} = (1+o(1))d^{-1}\trace G^{-1}$ (we have only removed one entry), and so we get 
\begin{eqnarray*}
    1-h_k = \frac{1}{1+c_k\,(1/d)\trace G^{-1}} + o(1).
\end{eqnarray*}

Now, by classical random matrix theory (RMT), it holds that $d^{-1}\trace (G-z\mathrm I_N)^{-1} \to m(z)$  
for any $z \in \mathbb C\setminus \mathbb R_+$, where the limiting Stieltjes transform $m$ satisfies the fixed-point equation
\begin{eqnarray}
    \frac{1}{m} = -z + \frac{1}{N}\sum_{k \in I}\frac{c_k}{1+c_k m}.
\end{eqnarray}
Defining $\kappa = \kappa(z) := 1/m(z)$, we have the equivalent formulation
\begin{eqnarray}
    \frac{1}{N}\sum_{k \in I}\frac{c_i \mu}{c_i + \mu} = \mu + z.
\end{eqnarray}
In particular since $N < S$, taking $z \to 0$ within $\mathbb C\setminus \mathbb R_+$, we get that $\mu$ satisfies the following fixed-point equation
\begin{eqnarray*}
    N = \sum_{k \in I}\frac{c_k}{c_k + \mu}.
\end{eqnarray*}
This gives, $h_k \simeq c_k/(c_k+\mu)$, $1-h_k \simeq \mu / (c_k + \mu)$ for all $k \in I$, and so
$$
\sum_{k \in I}c_k a_k\, (1-h_k) \simeq A \mu \sum_{k \in I} \frac{c_k}{c_k + \mu} = AN\mu = ANn\nu.
$$
which completes the proof of Proposition \ref{prop:main}. \qed

\subsection{Proof of Theorem \ref{thm:finite-n}}
Proposition \ref{prop:main} already separates the contributions of limited
representation and finite sampling. The remaining task is to replace
the empirical frequencies by their population counterparts and
understand how the resulting fixed point depends on the Zipf exponent.

To distinguish the two fixed points, write $\widehat\nu$ for the
sample-dependent quantity in Proposition~\ref{prop:main}, and let
$\nu>0$ solve
\begin{equation}
    N=\sum_{i=1}^d\frac{p_i}{p_i+\nu},
    \quad
    p_i=\frac{i^{-\alpha}}{H_{d,\alpha}}.
    \label{eq:proof-population-fp}
\end{equation}
For $0<N<d$, existence and uniqueness follow because the right-hand
side is continuous and strictly decreasing from $d$ to zero.
Throughout the proof, $\alpha$ is fixed, $d,N\to\infty$, and either
$N=o(d)$ or $N/d\to\psi_0\in(0,1)$.

We first analyze the population fixed point. Introduce
\[
    \psi:=\frac Nd,
    \quad
    \rho_d:=\frac{1}{H_{d,\alpha}d^\alpha\nu}.
\]
Then \eqref{eq:proof-population-fp} becomes
\[
    \psi
    =\frac1d\sum_{i=1}^d
        \frac{\rho_d}{\rho_d+(i/d)^\alpha}.
\]
The useful observation is that this sum is a Riemann approximation
to the scalar function
\[
    G_\alpha(\rho)
    :=\int_0^1\frac{\rho}{\rho+t^\alpha}\,\mathrm dt.
\]
Since the integrand is nonincreasing in $t$,
\begin{equation}
    0\le G_\alpha(\rho_d)-\psi\le\frac1d,
    \quad
    G_\alpha(\rho_d)
    =\psi\bigl(1+O(N^{-1})\bigr).
    \label{eq:proof-riemann}
\end{equation}
Meanwhile, the bias contribution has the exact representation
\begin{equation}
    B=AN\nu
     =A\frac{d^{1-\alpha}}{H_{d,\alpha}}
        \frac{\psi}{\rho_d}.
    \label{eq:proof-bias-rho}
\end{equation}
Thus all three scaling regimes follow from the behavior of
$G_\alpha$ and the normalization $H_{d,\alpha}$.

When the width is proportional to the vocabulary size,
$\psi\to\psi_0\in(0,1)$, continuity and strict monotonicity imply
\[
    \rho_d\sim\rho(\alpha,\psi),
    \quad
    G_\alpha\bigl(\rho(\alpha,\psi)\bigr)=\psi.
\]
For sublinear width, $\psi\to0$, and
\eqref{eq:proof-riemann} instead reduces the problem to the behavior
of $G_\alpha(\rho)$ as $\rho\downarrow0$.

For a subcritical exponent $0\le\alpha<1$, the function
$t^{-\alpha}$ is integrable on $(0,1)$. Dominated convergence
therefore gives
\[
    \frac{G_\alpha(\rho)}{\rho}
    =\int_0^1\frac{\mathrm dt}{\rho+t^\alpha}
    \longrightarrow\frac1{1-\alpha}.
\]
Together with
$H_{d,\alpha}\sim d^{1-\alpha}/(1-\alpha)$, this immediately
yields both subcritical laws. Indeed, if $N=o(d)$, then
$\rho_d\sim(1-\alpha)\psi$, and hence $B\sim A$.
For proportional width, the integral fixed point remains
nontrivial, giving
\[
    B\sim A(1-\alpha)\frac{\psi}{\rho(\alpha,\psi)}.
\]

At the critical exponent $\alpha=1$, the integral can be evaluated
explicitly:
\begin{equation}
    G_1(\rho)
    =\rho\log\left(1+\frac1\rho\right).
    \label{eq:proof-critical-integral}
\end{equation}
For $N=o(d)$, equations~\eqref{eq:proof-riemann}
and~\eqref{eq:proof-critical-integral} give
\[
    \psi\sim\rho_d\log(1/\rho_d).
\]
The logarithmic factor changes more slowly than $\rho_d$, so
\[
    \log(1/\psi)
    =\log(1/\rho_d)-\log\log(1/\rho_d)+o(1)
    \sim\log(1/\rho_d).
\]
Consequently,
\[
    \frac{\psi}{\rho_d}\sim\log(1/\psi)=\log(d/N).
\]
Using $H_{d,1}\sim\log d$ in \eqref{eq:proof-bias-rho},
we obtain
\[
    B\sim\frac{A}{\log d}\log\left(\frac dN\right).
\]

For proportional width, it is convenient to return to the discrete
fixed point and set
\[
    q:=\frac1{H_{d,1}\nu}.
\]
This gives
\[
    N=\sum_{i=1}^d\frac{q}{q+i},
    \quad
    B=\frac{AN}{H_{d,1}q}
      \sim\frac{A}{\log d}\frac Nq.
\]
In particular, the factor associated with this definition of $q$
is $N/q$. An explicit expression is obtained by writing
$b:=\psi/\rho$, where $\rho=\rho(1,\psi)$. The integral equation
then reads
\[
    b=\log\left(1+\frac b\psi\right),
\]
whose positive solution is
\[
    b=-\mathcal W_{-1}(-\psi e^{-\psi})-\psi.
\]
Thus, in the proportional-width regime,
\[
    B\sim\frac{A}{\log d}
    \left[-\mathcal W_{-1}(-\psi e^{-\psi})-\psi\right].
\]

The supercritical case $\alpha>1$ has a different small-$\rho$
behavior. Substituting $t=\rho^{1/\alpha}u$ gives
\[
    G_\alpha(\rho)
    =\rho^{1/\alpha}
      \int_0^{\rho^{-1/\alpha}}\frac{\mathrm du}{1+u^\alpha}.
\]
The integral now converges as $\rho\downarrow0$, with
\[
    I_\alpha
    :=\int_0^\infty\frac{\mathrm du}{1+u^\alpha}
    =\frac{\pi}{\alpha\sin(\pi/\alpha)}.
\]
Hence, for $N=o(d)$,
\[
    \psi\sim I_\alpha\rho_d^{1/\alpha},
    \quad
    \rho_d\sim\left(\frac{\psi}{I_\alpha}\right)^\alpha.
\]
Since $H_{d,\alpha}\to\zeta(\alpha)$,
equation~\eqref{eq:proof-bias-rho} becomes
\begin{align*}
    B
    &\sim
    \frac{A I_\alpha^\alpha}{\zeta(\alpha)}
        d^{1-\alpha}\psi^{1-\alpha}\\
    &=A C_\alpha N^{-(\alpha-1)},
    \quad
    C_\alpha:=
    \frac1{\zeta(\alpha)}
    \left[\frac{\pi}{\alpha\sin(\pi/\alpha)}\right]^\alpha.
\end{align*}
This constant applies to sublinear width. If $N/d\to\psi_0\in(0,1)$,
the vocabulary cutoff remains relevant, and the corresponding law is
\[
    B\sim
    \frac{A}{\zeta(\alpha)}
        d^{1-\alpha}\frac{\psi}{\rho(\alpha,\psi)}.
\]
Equivalently, the dependence is still proportional to
$N^{-(\alpha-1)}$, but its prefactor also depends on $\psi$.

It remains to justify replacing the empirical quantities in the
proposition by those analyzed above. The sample-size assumption
can be written as
\[
    \lambda_d:=np_d
    =\frac{n}{d^\alpha H_{d,\alpha}}
    \longrightarrow\infty.
\]
Thus even the least frequent context has a diverging expected
count. In particular,
\[
    \mathbb E[d-S]
    =\sum_{i=1}^d(1-p_i)^n
    \le\sum_{i=1}^d e^{-np_i}
    \le d e^{-\lambda_d}.
\]
Since $d-N\ge\delta d$ eventually for some $\delta>0$ in both
width regimes under consideration, the inequality
\[
    0\le(d-N)-(S-N)_+\le d-S
\]
implies
\begin{equation}
    \mathbb E\left|
        \frac{(S-N)_+}{d-N}-1
    \right|
    \le\frac{e^{-\lambda_d}}{\delta}
    \longrightarrow0.
    \label{eq:proof-occupancy}
\end{equation}
The deterministic variance contribution is therefore
\[
    V:=\frac{1-A}{n}(d-N)
      =(1-A)(1-\psi)\phi,
    \quad \phi:=\frac dn.
\]
For sublinear width, this reduces to
$V=(1-A)\phi(1+o(1))$.

To handle the fixed point, define
\[
    F(t):=\sum_{i=1}^d\frac{p_i}{p_i+t},
    \quad
    \widehat F(t):=
    \sum_{i=1}^d\frac{\widehat p_i}{\widehat p_i+t},
    \quad
    \widehat p_i:=\frac{c_i}{n}.
\]
Rather than requiring all empirical frequencies to concentrate
uniformly, it is enough to control this aggregate. For every $t>0$,
\[
    \left|
        \frac{\widehat p_i}{\widehat p_i+t}
        -\frac{p_i}{p_i+t}
    \right|
    \le
    \frac{p_i}{p_i+t}\,
    \frac{|\widehat p_i-p_i|}{p_i}.
\]
The binomial variance bound gives
\[
    \mathbb E\frac{|\widehat p_i-p_i|}{p_i}
    \le\frac1{\sqrt{np_i}}
    \le\frac1{\sqrt{\lambda_d}},
\]
and hence
\begin{equation}
    \mathbb E|\widehat F(t)-F(t)|
    \le\frac{F(t)}{\sqrt{\lambda_d}}.
    \label{eq:proof-fp-concentration}
\end{equation}

The population asymptotics also show that the root is stable under
these perturbations. For any fixed $a>0$,
\[
    \frac{F(a\nu)}{N}\longrightarrow
    \begin{cases}
        a^{-1},
        &N=o(d),\quad 0\le\alpha\le1,\\[3pt]
        a^{-1/\alpha},
        &N=o(d),\quad\alpha>1,\\[6pt]
        \displaystyle
        \frac{G_\alpha(\rho_0/a)}{G_\alpha(\rho_0)},
        &N/d\to\psi_0\in(0,1),
    \end{cases}
\]
where $G_\alpha(\rho_0)=\psi_0$.
Each limit is greater than one for $a<1$ and smaller than one
for $a>1$. For any fixed $\epsilon\in(0,1)$,
equation~\eqref{eq:proof-fp-concentration} therefore ensures that,
with probability tending to one,
\[
    \widehat F((1-\epsilon)\nu)>N,
    \quad
    \widehat F((1+\epsilon)\nu)<N.
\]
Monotonicity traps the empirical root between these two arguments:
\[
    (1-\epsilon)\nu\le\widehat\nu\le(1+\epsilon)\nu.
\]
Thus
\begin{equation}
    \widehat\nu/\nu\xrightarrow{\mathbb P}1.
    \label{eq:proof-root-concentration}
\end{equation}
Here, as in the proposition, $\widehat\nu=0$ when $N\ge S$;
the preceding argument also shows that this event has vanishing
probability in the regimes considered.

For an expectation over the sample, we need slightly more than
convergence in probability. A useful feature of the fixed point
is that its solution is concave in the frequency vector.
Indeed, let $\nu_N(u)$ denote the solution for a nonnegative
vector $u$, setting it to zero when $u$ has at most $N$ nonzero
entries. For $t\ge0$,
\[
    t\le\nu_N(u)
    \quad\Longleftrightarrow\quad
    \sum_{i=1}^d\frac{u_i t}{u_i+t}-Nt\ge0,
\]
with the continuous extension at $u_i=t=0$.
Because $u_i t/(u_i+t)$ is jointly concave, this describes a
convex hypograph. Jensen's inequality consequently gives
\[
    \mathbb E[\widehat\nu]
    =\mathbb E[\nu_N(\widehat p)]
    \le\nu_N(\mathbb E[\widehat p])
    =\nu.
\]
Together with \eqref{eq:proof-root-concentration}, this implies
relative $L^1$ convergence. To see this, put
$Z_d:=\widehat\nu/\nu$. Then $Z_d\ge0$, $\mathbb E[Z_d]\le1$,
and $Z_d\to1$ in probability, so
\[
    \mathbb E|Z_d-1|
    =\mathbb E[Z_d]-1+2\mathbb E[(1-Z_d)_+]
    \le2\mathbb E[(1-Z_d)_+]
    \longrightarrow0.
\]

We can now return to Proposition~\ref{prop:main}. Writing its
remainder as $\varepsilon(c)$, the conditional expected regret is
\[
    \mathbb E_{Y,\Omega}[R_{n,d}(N)\mid c]
    =
    AN\widehat\nu+\frac{1-A}{n}(S-N)_+
    +\varepsilon(c),
\]
where the expectation includes the next-token observations unless
they have already been averaged in the definition of the regret.
The estimates above show that the two displayed main terms equal
\[
    B+V+o_{L^1}(B+V).
\]
Therefore, provided the proposition's approximation error satisfies
$\mathbb E|\varepsilon(c)|=o(B+V)$, we conclude that
\[
    \mathbb E_{X,Y,\Omega}[R_{n,d}(N)]
    \sim B+V.
\]
The three forms of $B$ follow from the population fixed-point
analysis, and the stated variance terms follow from
\eqref{eq:proof-occupancy}.

The restriction $\psi_0<1$ is important: when $N/d\to1$, the
quantity $d-N$ may be too small for the occupancy estimate above
to give a relative approximation. At $N=d$, however, the separate
interpolation branch applies directly: $\nu=0$, and both
contributions vanish. \qed